# Super-Mixed Multiple Attribute Group Decision Making Method Based on Hybrid Fuzzy Grey Relation Approach Degree[*]

Gol Kim [a], Fei Ye [b]

[a] Center of Natural Science, University of Sciences, Pyongyang, DPR Korea
[b] School of Business Administration, South China University of Technology, Guangzhou 510640, China (yefei@scut.edu.cn)

**Abstract** —The feature of our method different from other fuzzy grey relation method for super-mixed multiple attribute group decision-making is that all of the subjective and objective weights are obtained by interval grey number and that the group decision-making is performed based on the relative approach degree of grey TOPSIS, the relative approach degree of grey incidence and the relative membership degree of grey incidence using 4-dimensional Euclidean distance. The weighted Borda method is used to obtain final rank by using the results of four methods. An example shows the applicability of the proposed approach.

**Keywords:** Grey fuzzy decision making, Super-mixed multiple attribute , Grey incidence degree, 4-dimensional Euclidean distance

## 1. Introduction

A multiple attribute decision making (MADM), in which attributes are real number, interval real number, linguistic and uncertain linguistic value, has been already applied in practice such as the evaluation of enterprise effect, the selection of investment project, the selection of person, the research of military equipment scheme, the evaluation of strategy effect, the reliability assessment and the maintainability assessment, etc (Yongqi Xia , 2004 , Dang Luo, Sifeng Liu , 2005, Yongqing Wei, Peide Liu , 2009).
Extended TOPSIS Method with Interval-Valued Intuitionistic Fuzzy Numbers for Virtual Enterprise Partner Selection has been researched by Fei Ye(2010).
 Chuanming Ding (2007,a) defined a new similarity degree for various types of attribute and normalized the calculation of similarity degree of the attribute value of each type in unified metric space. Also, by this similarity degree, the comparison of each plan with ideal plan was performed and decision making method was given.  Chuanming (2007,b), based on the TOPSIS (Technique for Order Preference by Similarity to Ideal Solution), transformed the attribute value of plan into four-dimensional attribute value, unified various types of attribute value, defined a four-dimensional approach degree, and by this approach degree, solved the multiple attribute mixed-type decision-making problem associated with real number, interval real number, linguistic and uncertain linguistic value. Yongqi Xia (2004) studied a method considering insufficiency degree of information and preference to danger on the basis of the grey-fuzzy comprehensive evaluation method of interval value preference. In the method, they represent the weight and the attribute value by two interval number pair by considering membership and grey degree at the same time. Sifeng Liu, Yaoguo Dang, Jiangling Wang, Zhengpeng Wu (2009), based on the definitions of entropy, proposed a method of getting weight that considers the character of grey cluster decision-making and 2-tuple linguistic assessment, and proposed the method of 2-tuple linguistic assessment based on grey cluster. Zhen Zhang, Chonghui Guo (2012) transformed  uncertain linguistic evaluation information of each decision maker to trapezoidal fuzzy numbers, and then denoted, by solving two optimization models, the collective evaluation of the alternatives by trapezoidal fuzzy numbers.

---

[*] This paper was supported in part by *South China University of Technology*, China

Yongqing Wei, Peide Liu (2009) constructed a evaluation indicator's system and the evaluation procedures based on the uncertain linguistic and TOPSIS method. Peide Liu, Yu Su (2010) introduced a concept of the trapezoid fuzzy linguistic variables, and defined the distance between two trapezoid fuzzy linguistic variables. They determined the combined weights of each attribute by the maximal deviation method and the nonlinear weighted comprehensive method, and defined a relative closeness degree to determine the ranking order of all alternatives by calculating the distances to both the positive ideal solution and negative ideal solution, respectively. Alecos M. Kelemenis, D. Th. Askounis (2009) proposed a multi-criteria approach to deal with group decision making under fuzzy environment and introduced a new reference point apart from the positive ideal solution and the negative ideal solution. Ning-ning Zhu, Jian-jun Zhu, Yen Ding (2009) studied the MADM problem of uncertain three-point linguistic information and obtained the weight vector of indicators by linear goal programming. They proposed the calculation steps of MADM problem of uncertain three-point linguistic information. Ting-yu Chen (2011), by extending the TOPSIS, proposed a useful method based on generalized interval valued trapezoidal fuzzy numbers (GITrFNs) for solving multiple criteria decision analysis (MCDA) problems, which employed the concept of signed distances to establish a simple and effective MCDA method based on the main structure of TOPSIS.

In this paper, we propose a generalized fuzzy grey decision making method taking into consideration of the grey degree of the weight and the attribute value at the same time, where attributes have the generalized super mixed-type values given by real number, interval value, linguistic value and uncertain linguistic value. First, we obtain the grey degree corresponding to fuzzy part of the grey fuzzy comprehensive decision matrix, and then obtain, by using the information sufficiency degree, a generalized decision matrix composed of four-dimensional vector dealing with the grey part of decision matrix. Second, we obtain four ranks for the generalized decision matrix composed of four-dimensional vector by four methods. Finally, using the ranks obtained from the above four methods, the final rank are determined by the weighted Borda method. An example is given to show the advantage of our method.

## 2. The formulation of decision making problem with super-mixed multiple attributes

If the attributes of same class include the values of each other different type, then its attributes are called super multiple attribute mixed type.

Let $A = \{A_1, A_2, \cdots, A_n\}$ be a set of plans, $G = \{G_1, G_2, \cdots, G_m\}$ be a set of attributes and $T = \{T_1, T_2, T_3, T_4\}$ = {real, interval real, linguistic, uncertain linguistic}
be a set of attribute type (the concrete definition is given below).

[**Definition 1**] Let $\tilde{\mu} = S(a)$ be linguistic value, where $S(a)$ is a linguistic measure. They are given by
$S = \{S(-5), S(-4), \cdots, S(4), S(5)\}$ = { extremely low, very low, low, comparatively low, a little low, general, a little high, comparatively high, high, very high, extremely high}, $a = \{-5, -4, \cdots, 4, 5\}$.
Supposing that $S(-5) \prec S(-4) \prec \cdots \prec S(4) \prec S(5)$, if $S(i_1) \prec S(i_2) \prec \cdots \prec S(i_n)$, then $\max\{S(i_1), S(i_2), \cdots, S(i_n)\} = S(i_n)$ and $\min\{S(i_1), S(i_2), \cdots, S(i_n)\} = S(i_1)$.

Each linguistic value can be represented by triangle fuzzy number $S = [a^L, a^M, a^U]$, $a^L \leq a^M \leq a^U$ and its membership function is given by

$$\mu_S(x) = \begin{cases} (x - a^L)/(a^M - a^L), & a^L \leq x \leq a^M \\ (x - a^L)/(a^M - a^L), & a^L \leq a^M \leq a^L \\ 0, & \text{otherwise} \end{cases}.$$

The expression forms of triangle fuzzy number corresponding to $S$ are given as follows.

'extremely low' = [0, 0, 0.1], 'very low'= [0, 0.1, 0.2], 'low'= [0.1, 0.2, 0.3], 'comparatively low'= [0.2, 0.3, 0.4], 'a little low' = [0.3, 0.4, 0.5], 'ordinary' = [0.4, 0.5, 0.6], 'a little high'=[0.5, 0.6, 0.7], 'comparatively high' = [0.6, 0.7, 0.8], 'high' = [0.7, 0.8, 0.9], 'very high'= [0.8, 0.9, 1.0], 'extremely high' = [0.9, 1.0, 1.0].

[**Definition 2**] Let $\tilde{A} = [\alpha, \beta, \gamma, \delta]$ be trapezoid fuzzy number. Then, its membership function is defined by

$$\mu_{\tilde{A}}(x) = \begin{cases} (x-\alpha)/(\beta-\alpha), & \alpha \leq x \leq \beta \\ 1, & \beta < x < \gamma \\ (\delta-x)/(\delta-\gamma), & \gamma \leq x \leq \delta \\ 0, & \text{otherwise} \end{cases}.$$

Let $\tilde{A} = [\alpha_1, \beta_1, \gamma_1, \delta_1]$ and $\tilde{B} = [\alpha_2, \beta_2, \gamma_2, \delta_2]$ be trapezoid fuzzy numbers, respectively. Then, the operational laws of trapezoid fuzzy number are as follows.

$$\tilde{A} \oplus \tilde{B} = [\alpha_1+\alpha_2, \beta_1+\beta_2, \gamma_1+\gamma_2, \delta_1+\delta_2],$$
$$\tilde{A} \otimes \tilde{B} = [\alpha_1\alpha_2, \beta_1\beta_2, \gamma_1\gamma_2, \delta_1\delta_2], k\tilde{A} \otimes \tilde{B} = [k\alpha_1, k\beta_1, k\gamma_1, k\delta_1].$$

[**Definition 3**] Let $S^L = [a^L, a^M, a^U]$ and $S^U = [b^L, b^M, b^U]$. A trapezoid fuzzy number $\tilde{\mu} = (a^L, a^M, b^M, b^U)$ defined by the membership function such as

$$\mu(x) = \begin{cases} (x-a^L)/(a^M-a^L), & a^L \leq x \leq a^M \\ 1, & a^M \leq x \leq b^M \\ (x-b^U)/(b^M-b^L), & b^M \leq x \leq b^U \\ 0, & \text{otherwise} \end{cases}$$

is called a uncertain linguistic value with lower bound $S^L$ and upper bound $S^U$.

[**Definition 4**] Let $a_{ij} = (a_{ij}^{(1)}, a_{ij}^{(2)}, a_{ij}^{(3)}, a_{ij}^{(4)})$ and $a_{ij}^{(1)} \leq a_{ij}^{(2)} \leq a_{ij}^{(3)} \leq a_{ij}^{(4)}$. Then, $a_{ij}$ is called a generalized attribute value of $i$ th plan for attribute $j$.

The concrete types of $a = (a^{(1)}, a^{(2)}, a^{(3)}, a^{(4)})$ are such as
- real number type: $a^{(1)} = a^{(2)} = a^{(3)} = a^{(4)}$;
- interval real number type: $a^{(1)} = a^{(2)} < a^{(3)} = a^{(4)}$;
- linguistic value type: $a^{(1)} < a^{(2)} = a^{(3)} < a^{(4)}$;
- uncertain linguistic value type: $a^{(1)} < a^{(2)} = a^{(3)} < a^{(4)}$.

[**Definition 5**] The decision making of mixed type multi attribute which attribute values of each other different attribute type among the same attribute are included is called super- mixed multi attribute decision making.

[**Definition 6**] Let $a = (a^{(1)}, a^{(2)}, a^{(3)}, a^{(4)})$ and $b = (b^{(1)}, b^{(2)}, b^{(3)}, b^{(4)})$ be the generalized attribute values, respectively. A distance of between $a$ and $b$ is defined by

$$d(a,b) = \sqrt{(b^{(1)}-a^{(1)})^2 + (b^{(2)}-a^{(2)})^2 + (b^{(3)}-a^{(3)})^2 + (b^{(4)}-a^{(4)})^2}$$

Let $a_i = \{a_{i1}, a_{i2}, \cdots, a_{im}\}$ be an attribute vector of $i$ plan and $\tilde{R}_{\otimes} = \{(a_{ij}, [v_{ij}^-, v_{ij}^+])\}_{n \times m}$ be a decision matrix, where $[v_{ij}^-, v_{ij}^+]$ is an interval grey number representing a grey degree of $a_{ij}$. Then a normalized decision matrix $\tilde{X}_{\otimes}$ is obtained by $\tilde{X}_{\otimes} = \{(x_{ij}, [v_{ij}^-, v_{ij}^+])\}_{n \times m}$.

If $a_{qr}$ is cost-type attribute, interval number $x_{qr} = [\underline{x}_{qr}, \bar{x}_{qr}]$ is obtained by normalization such as

$$\underline{x}_{qr} = \frac{1/\bar{a}_{qr}}{\sum_{q=1}^{n}(1/\underline{a}_{qr})}, \quad \bar{x}_{qr} = \frac{1/\underline{a}_{qr}}{\sum_{q=1}^{n}(1/\bar{a}_{qr})}$$

and if $a_{qr}$ is effect-type attribute, interval number $x_{qr}$ is obtained by normalization such as $\underline{x}_{qr} = \dfrac{\underline{a}_{qr}}{\sum_{q=1}^{n} \overline{a}_{qr}}$, $\overline{x}_{qr} = \dfrac{\overline{a}_{qr}}{\sum_{q=1}^{n} \underline{a}_{qr}}$. If $a_{qr} = [a_{qr}^L, a_{qr}^*, a_{qr}^U]$, then $x_{qr} = [x_{qr}^L, x_{qr}^*, x_{qr}^U]$ is obtained by the normalization such as $x_{qr}^L = \dfrac{a_{qr}^L}{\sum_{q=1}^{n} a_{qr}^*}$, $x_{qr}^* = \dfrac{a_{qr}^*}{\sum_{q=1}^{n} a_{qr}^*}$, $x_{qr}^U = \dfrac{a_{qr}^U}{\sum_{q=1}^{n} a_{qr}^*}$.

If $a_{qr} = [a_{qr}^L, a_{qr}^*, a_{qr}^{**}, a_{qr}^U]$, $x_{qr} = [x_{qr}^L, x_{qr}^*, x_{qr}^{**}, x_{qr}^U]$ is obtained by normalization such as $x_{qr}^L = \dfrac{a_{qr}^L}{\sum_{q=1}^{n} a_{qr}^*}$,

$x_{qr}^* = \dfrac{a_{qr}^*}{\sum_{q=1}^{n} a_{qr}^*}$, $x_{qr}^{**} = \dfrac{a_{qr}^{**}}{\sum_{q=1}^{n} a_{qr}^{**}}$, $x_{qr}^U = \dfrac{a_{qr}^U}{\sum_{q=1}^{n} a_{qr}^{**}}$.

## 3. Determining of attribute weights

### 3.1. Subjective weight of attributes

Let $\alpha_l = [\alpha_l^1, \cdots, \alpha_l^j, \cdots, \alpha_l^m]$, ($l = \overline{1, L}$) be the attribute weights determined by AHP from the decision-making group. The weight of attribute $G_j$ is given as interval grey number $\alpha_j(\otimes) \in [\underline{\alpha}_j, \overline{\alpha}_j]$, where $\underline{\alpha}_j = \min_{1 \le l \le L}\{\alpha_l^j\}$, $\overline{\alpha}_j = \max_{1 \le l \le L}\{\alpha_l^j\}$.

### 3.2. Objective weight of attributes

#### 3.2.1 Objective weight by optimization

We define the deviation of decision plan $A_i$ from all other decision plans for attribute $G_j$ in normalized decision matrix $X = (x_{ij}(\otimes))_{n \times m}$ as follows.

$$D_{ij}(\beta^{opt}) = \sum_{k=1}^{m} d(x_{ij}, x_{kj})\beta_j^{opt} =$$

$$= \sum_{k=1}^{m} \sqrt{(x_{kj}^{(1)} - x_{ij}^{(1)})^2 + (x_{kj}^{(2)} - x_{ij}^{(2)})^2 + (x_{kj}^{(3)} - x_{ij}^{(3)})^2 + (x_{kj}^{(4)} - x_{ij}^{(4)})^2} \beta_j^{opt}.$$

In order to choose a proper weight vector $\beta^{opt}$ such that sum of overall deviation for the decision plan attains maximum, we define a deviation function such as

$$D(\beta) = \sum_{j=1}^{m}\sum_{i=1}^{n}\sum_{k=1}^{n} d(x_{ij}, x_{kj})\beta_j$$

and solve the following nonlinear programming problem.

[P1]   $\max D(\beta) = \sum_{j=1}^{m}\sum_{i=1}^{n}\sum_{k=1}^{n} d(x_{ij}, x_{kj})\beta_j$,

$\text{s.t.} \sum_{j=1}^{m} \beta_j^2 = 1$, $\beta_j \ge 0$, $j = \overline{1, m}$

[Theorem 1] The solution of problem P1 is given by

$$\overline{\beta}_j = \dfrac{\sum_{i=1}^{n}\sum_{k=1}^{n} d(x_{ij}, x_{kj})}{\sqrt{\sum_{j=1}^{m}\left[\sum_{i=1}^{n}\sum_{k=1}^{n} d(x_{ij}, x_{kj})\right]^2}}, j = \overline{1, m}$$

By the normalization of $\overline{\beta}_j$, $j = \overline{1,m}$, we obtain

$$\beta_j^{opt} = \frac{\sum_{i=1}^{n}\sum_{k=1}^{n} d(x_{ij}, x_{kj})}{\sum_{j=1}^{m}\sum_{i=1}^{n}\sum_{k=1}^{n} d(x_{ij}, x_{kj})}, \quad j = \overline{1,m}$$

**3.2.2. Objective weight by entropy method**

The entropy weights of the generalized attribute value $x_{ij} = (x_{ij}^{(1)}, x_{ij}^{(2)}, x_{ij}^{(3)}, x_{ij}^{(4)})$ are obtained for each $x_{ij}^{(k)}$ ($k = 1,2,3,4$) as follows. The value $x_{ij}^{(k)}$ ($k = 1,2,3,4$) is normalized by $p_{ij}^{(k)} = \frac{x_{ij}^{(k)}}{\sum_{i=1}^{n} x_{ij}^{(k)}}$ ($i = \overline{1,n}$, $j = \overline{1,m}$). The entropy value of the $j$ th attribute is $E_j^{(k)} = -\frac{1}{\ln n}\sum_{i=1}^{n} p_{ij}^{(k)} \ln p_{ij}^{(k)}$ ($j = \overline{1,m}$). In the above formula, if $p_{ij}^{(k)} = 0$, then we put that $p_{ij}^{(k)} \ln p_{ij}^{(k)} = 0$. Then deviation coefficient for the $j$ th attribute is calculated by $\eta_j^{(k)} = 1 - E_j^{(k)}$ ($j = \overline{1,m}$).

Thus, the entropy weight $\beta^{(k)ent} = (\beta_1^{(k)ent}, \beta_2^{(k)ent}, \cdots, \beta_j^{(k)ent}, \cdots, \beta_m^{(k)ent})$ for the component $x_{ij}^{(k)}$ ($k = 1,2,3,4$) is such as

$$\beta_j^{(k)ent} = \frac{\eta_j^{(k)}}{\sum_{j=1}^{m} \eta_j^{(k)}} = \frac{1 - E_j^{(k)}}{\sum_{j=1}^{m}(1 - E_j^{(k)})} = \frac{1 - E_j^{(k)}}{m - \sum_{j=1}^{m} E_j^{(k)}} \quad (j = \overline{1,m}, k = \overline{1,4}).$$

**3.2.3. Determining of comprehensive objective weights**

The comprehensive objective weight is determined by the interval grey number
$\beta(\otimes) = (\beta_1(\otimes), \beta_2(\otimes), \cdots, \beta_j(\otimes), \cdots, \beta_m(\otimes))$, $\beta_j(\otimes) \in [\underline{\beta}_j, \overline{\beta}_j]$

$$\underline{\beta}_j(\otimes) = \min\{\beta_j^{opt}, \beta_j^{(1)ent}, \beta_j^{(2)ent}, \beta_j^{(3)ent}, \beta_j^{(4)ent}\},$$

$$\overline{\beta}_j(\otimes) = \max\{\beta_j^{opt}, \beta_j^{(1)ent}, \beta_j^{(2)ent}, \beta_j^{(3)ent}, \beta_j^{(4)ent}\}.$$

**3.3. Determining of final comprehensive weights**

The final comprehensive weight is determined by

$$w_j(\otimes) = \frac{\alpha_j(\otimes) \times \beta_j(\otimes)}{\sum_{j=1}^{m} \alpha_j(\otimes) \times \beta_j(\otimes)}, \quad j = \overline{1,m}$$

where $\alpha_j(\otimes)$ and $\beta_j(\otimes)$ are the subjective weight and the objective weight for $j$ th attribute, respectively. Thus, the weight of the attribute $G_j$ is given by the interval grey number $w_j(\otimes) \in [\underline{w}_j, \overline{w}_j]$, $0 \le \underline{w}_j \le \overline{w}_j \le 1$, $j = \overline{1,m}$.

## 4. Evaluation methods for decision making plans

In the case which a grey degree of attribute weight value is also given at the same time, the comprehensive weights is represented by the grey-fuzzy number such as

$$\widetilde{B}_{\otimes} = [([\underline{w}_1, \overline{w}_1], [\underline{r}_1, \overline{r}_1]), ([\underline{w}_2, \overline{w}_2], [\underline{r}_2, \overline{r}_2]), \cdots, ([\underline{w}_m, \overline{w}_{m1}], [\underline{r}_m, \overline{r}_m])],$$

where $[\underline{r}_i, \overline{r}_i]$ is an grey interval number representing the grey degree of the grey-fuzzy weight $w_i$. By normalizing $\widetilde{B}_{\otimes}$, we have the weight vector such as

$$\widetilde{W}_\otimes = [(\widetilde{w}_1(\otimes), s_1(\otimes)), \cdots, (\widetilde{w}_m(\otimes), s_m(\otimes))], \ (\widetilde{w}_j(\otimes), s_j(\otimes)) = ([w_j^-, w_j^+], [s_j^-, s_j^+])$$

where $w_j^- = \dfrac{\underline{w}_j}{\sum_{i=1}^m \overline{w}_j}$, $w_j^+ = \dfrac{\overline{w}_j}{\sum_{i=1}^m \underline{w}_j}$, $s_j^- = \dfrac{\underline{r}_j}{\sum_{i=1}^m \overline{r}_j}$, $s_j^+ = \dfrac{\overline{r}_j}{\sum_{i=1}^m \underline{r}_j}$, $j = \overline{1, m}$

## 4.1. Evaluation of plan by the relative approach degree of grey TOPSIS method

Assume that the subjective preference value of the plan $A_i$ is given by the generalized value $q_i = (q_i^{(1)}, q_i^{(2)}, q_i^{(3)}, q_i^{(4)})$. Let $\widetilde{Z} = \{z_{ij}\}_{n \times m}$ be the normalized decision matrix with the subjective preference such as

$$z_{ij} = \left(\dfrac{1}{2}(q_i^{(1)} + x_{ij}^{(1)}), \dfrac{1}{2}(q_i^{(2)} + x_{ij}^{(2)}), \dfrac{1}{2}(q_i^{(3)} + x_{ij}^{(3)}), \dfrac{1}{2}(q_i^{(4)} + x_{ij}^{(4)})\right).$$

Then, the normalized grey-fuzzy decision matrix with the subjective preference is given by $\widetilde{Z}_\otimes = \{z_{ij}, [\mu_{ij}^-, \mu_{ij}^+]\}_{n \times m}$ and the comprehensive weighted decision matrix is given by $\widetilde{Y}_\otimes = \widetilde{W}_\otimes \circ \widetilde{Z}_\otimes = [\widetilde{y}_{ij}]_{\otimes n \times m} = [(\widetilde{y}_{ij}, y_{ij})_\otimes]_{n \times m}$, where $\widetilde{y}_{ij} = \widetilde{w}_j(\otimes) z_{ij} = (\widetilde{y}_{ij}^{(1)}, \widetilde{y}_{ij}^{(2)}, \widetilde{y}_{ij}^{(3)}, \widetilde{y}_{ij}^{(4)})$ and $y_{ij} = \mu_{ij}(\otimes) \circ s_{ij}(\otimes) = [t_{ij}^-, t_{ij}^+]$ is the fuzzy part and grey part of $\widetilde{y}_{ij}$, respectively. Thus, we obtain the decision matrix $\widetilde{Y}_\otimes = \{(\widetilde{y}_{ij}, [t_{ij}^-, t_{ij}^+])\}_{n \times m}$.

The grey part is processed by a danger index as follows. For the interval grey number $[t_{ij}^-, t_{ij}^+] \subset [0,1]$, let $M = \dfrac{1}{2}(t_{ij}^- + t_{ij}^+)$, $D = \dfrac{1}{2}(t_{ij}^+ - t_{ij}^-)$, $F_{ij}(\alpha) = M + (2\alpha - 1)D$, $\alpha \in [0,1]$, and

$$G_{ij}(\alpha) = 1 - F_{ij}(\alpha) = \begin{cases} 1 - t_{ij}^-, & \alpha = 0 \\ 1 - t_{ij}^+, & \alpha = 1 \\ 1 - \dfrac{1}{2}(t_{ij}^- + t_{ij}^+), & \alpha = \dfrac{1}{2} \end{cases}$$

where $\alpha$ is called a danger index. Then $G_{ij}(\alpha)$ is a conversion formula of information sufficiency degree.

Given the danger index $\alpha$ of the decision maker, we obtain the final decision matrix $Y = (y_{ij})_{n \times m}$, where $y_{ij} = (y_{ij}^{(1)}, y_{ij}^{(2)}, y_{ij}^{(3)}, y_{ij}^{(4)})$ and $y_{ij}^{(k)} = \widetilde{y}_{ij}^{(1)} G_{ij}(\alpha)$, $(k = \overline{1,4})$. The attribute vector of each plan for the comprehensive weighted decision 
matrix is given by $y_i = \{y_{i1}, y_{i2}, \cdots, y_{ij}, \cdots, y_{im}\}$ $(i = \overline{1, n})$.

**[Definition 6]** Let $y_j^+ = \{y_j^{+(1)}, y_j^{+(2)}, y_j^{+(3)}, y_j^{+(4)}\}$ and $y_j^{+(k)} = \max_{1 \leq i \leq n}\{y_{ij}^{(k)}\}$ $(k = 1,2,3,4)$.

A $m$-dimensional interval grey number vector $y^+ = \{y_1^+, y_2^+, \cdots, y_m^+\}$ is called a positive ideal attribute plan vector.

Let $y_j^- = (y_j^{-(1)}, y_j^{-(2)}, y_j^{-(3)}, y_j^{-(4)})$ and $y_j^{-(k)} = \min_{1 \leq i \leq n}\{y_{ij}^{(k)}\}$ $(k = 1,2,3,4)$. Then $y^- = \{y_1^-, y_2^-, \cdots, y_m^-\}$ is called a negative ideal attribute plan vector.

Euclidian distance between each plan attribute vector $y_i$ and the positive or negative ideal plan attribute vector $y^+$ or $y^-$ is

$$D_i^+ = \sqrt{\sum_{j=1}^m \left[(y_{ij}^{(1)} - y_j^{+(1)})^2 + (y_{ij}^{(2)} - y_j^{+(2)})^2 + (y_{ij}^{(3)} - y_j^{+(3)})^2 + (y_{ij}^{(4)} - y_j^{+(4)})^2\right]}$$

or
$$D_i^- = \sqrt{\sum_{j=1}^{m}\left[(y_{ij}^{(1)} - y_j^{-(1)})^2 + (y_{ij}^{(2)} - y_j^{-(2)})^2 + (y_{ij}^{(3)} - y_j^{-(3)})^2 + (y_{ij}^{(4)} - y_j^{-(4)})^2\right]}\ .$$

The relative approach degree between each evaluation plan and the ideal plan is
$$C_i = \frac{D_i^-}{D_i^+ + D_i^-},\ i = \overline{1,n}.$$

The best plan is one corresponding to the largest $C_i$.

## 4.2. Evaluation of plan by the relative approach degree of grey incidence

**[Definition 7]** Let $Y = \{y_{ij}\}_{n \times m}$ be the normalized comprehensive weighted decision matrix and $y_j^+$ and $y_j^-$ be the positive and negative ideal plan attribute vector, respectively. We define
$$r_{ij}^+ = \frac{\min_i \min_j d(y_{ij}, y_j^+) + \rho \max_i \max_j d(y_{ij}, y_j^+)}{d(y_{ij}, y_j^+) + \max_i \max_j d(y_{ij}, y_j^+)},$$
$$r_{ij}^- = \frac{\min_i \min_j d(y_{ij}, y_j^-) + \rho \max_i \max_j d(y_{ij}, y_j^-)}{d(y_{ij}, y_j^-) + \max_i \max_j d(y_{ij}, y_j^-)}.$$

Then, $r_{ij}^+$ ($r_{ij}^-$) is called the coefficient of positive (negative) ideal grey interval incidence with respect to the positive ideal attribute value $y_j^+$ ($y_j^-$), where $\rho \in (0,1)$ and generally $\rho = 0.5$ is taken.

**[Definition 8]** The matrix $P^+ = \{r_{ij}^+\}_{n \times m}$ ($P^- = \{r_{ij}^-\}_{n \times m}$) is called a grey incidence coefficient matrix of the given plan with respect to the positive (negative) ideal plan.

**[Definition 9]** Let $G(y^+, y_i) = \frac{1}{m}\sum_{j=1}^{m} r_{ij}^+$, $G(y^-, y_i) = \frac{1}{m}\sum_{j=1}^{m} r_{ij}^-$, $i = \overline{1,n}$.

Then $G(y^+, y_i)$ ($G(y^-, y_i)$) is called a degree of grey interval incidence of the comprehensive attribute vector for the plan $A_i$ with respect to the positive (negative) ideal plan attribute vector.

**[Theorem 2]** The grey interval incidence degrees $G(y^+, y_i)$ and $G(y^-, y_i)$ satisfy the four axioms of grey incidence degree, i.e. normality, pair-symmetry, wholeness and closeness.

The degree of grey incidence relative approach is defined by introducing the preference coefficients as follows.
$$C_i = \begin{cases} \dfrac{G(y^+, y_i) \cdot \theta_+}{G(y^+, y_i) \cdot \theta_+ + G(y^-, y_i) \cdot \theta_-}; & 0 < \theta_+ < 1,\ \theta_- < 1 \\ G(y^+, y_i); & \theta_+ = 1,\ \theta_- = 0 \end{cases},$$

where $\theta_+$ and $\theta_-$ are the preference coefficients, respectively. Generally, we regard as $\theta_+ > \theta_-$ and choose it so as to satisfy $0 < \theta_+ \leq 1$, $0 < \theta_- \leq 1$, $\theta_+ + \theta_- = 1$. The optimal plan corresponds to the largest value among of the relative approach degree $C_i$.

## 4.3. Evaluation of plan by the relative membership degree of grey incidence

If the membership degree of the positive ideal plan with respect to the plan $A_i$ is $u_i$, the membership degree of the negative ideal plan corresponding to the plan $A_i$ is $1 - u_i$. Therefore, we can find the membership degree vector $u = (u_1, u_2, \cdots, u_n)$ by solving the following problem.

[P2]    $\min F(u) = \sum_{i=1}^{n}\left\{\left[(1 - u_i)G(y^+, y_i)\right]^2 + \left[u_i G(y^-, y_i)\right]^2\right\}.$

**[Theorem 3]** The optimal solution of the optimization problem P2 is given by

$$u_i = \frac{G^2(y^+, y_i)}{G^2(y^+, y_i) + G^2(y^-, y_i)}, (i = \overline{1,n})$$

The optimal plan is one having the largest membership degree $u_i$.

## 4.4. Evaluation of plan by the grey relation relative approach degree using maximum entropy estimation

**[Definition 10]** Let $G(y^+, y_i)$ and $G(y^-, y_i)$ be the grey interval incidence degrees for the plan $A_i$ with respect to the positive ideal plan and the negative ideal plan, respectively. We denote the weights of these two grey interval incidence degrees by $\beta_1$ and $\beta_2$ ($\beta_1 + \beta_2 = 1$, $\beta_1, \beta_2 \geq 0$), respectively. Then,

$$C_i'' = \beta_1 G(y^+, y_i) + \beta_2 [1 - G(y^-, y_i)] \ (i = \overline{1,n})$$

is called a grey comprehensive incidence degree of the factor vector $y_i$.

We determine $\beta_1$ and $\beta_2$ by entropy method. Thus, we solve the following optimization problem

[P3] $\quad \max\{\sum_{i=1}^{n}[\beta_1 G(y^+, y_i) + \beta_2(1 - G(y^-, y_i))] - \sum_{j=1}^{2}\beta_j \ln \beta_j\}$

$$s.t. \begin{cases} \beta_1 + \beta_2 = 1, \\ \beta_1 \geq 0, \beta_2 \geq 0 \end{cases}.$$

The solution of [P3] is such as

$$\beta_1 = e^{\sum_{i=1}^{n}(G(y^+, y_i) + G(y^-, y_i) - 1)} (1 + e^{\sum_{i=1}^{n}(G(y^+, y_i) + G(y^-, y_i))})^{-1},$$

$$\beta_2 = (1 + e^{\sum_{i=1}^{n}(G(y^+, y_i) + G(y^-, y_i))})^{-1}.$$

The best plan is one having the largest value of $C_i^*$.

## 4.5. Final rank methods for decision making plans

The final rank is determined by the weighted Borda method using rank vectors obtained from the above four methods.

## 5. An illustrative example

Let's consider the decision-making problem for the fighter development plan of some types. The decision matrix is given by the super multiple attribute mixed type in Table 1. The meaning of attributes is such as; $G_1$ - weight empty of body(Kg), $G_2$ - flight radius(Km), $G_3$ - maximum flying speed(Km/h), $G_4$ - development cost (ten thousand Yuan), $G_5$ - reversal of body head(h), $G_6$ - maintenance possibility, $G_7$ - security, $G_8$ - reliability level of development group, $G_9$ - degree of environmental influence.

Assume that two experts are invited to determine the subjective attribute weights by AHP method. Thus, the subjective weight obtained from group AHP method is given by the grey number such as
$\alpha(\otimes) = ([\underline{\alpha}_1, \overline{\alpha}_1], [\underline{\alpha}_2, \overline{\alpha}_2], [\underline{\alpha}_3, \overline{\alpha}_3], [\underline{\alpha}_4, \overline{\alpha}_4], [\underline{\alpha}_5, \overline{\alpha}_5], [\underline{\alpha}_6, \overline{\alpha}_6], [\underline{\alpha}_7, \overline{\alpha}_7], [\underline{\alpha}_8, \overline{\alpha}_8], [\underline{\alpha}_9, \overline{\alpha}_9])$
= ([0.2305, 0.3093], [0.1501, 0.1675], [0.1262, 0.1761], [0.1323, 0.1348], [0.0815, 0.0948], [0.0557, 0.0622], [0.0431, 0.0623], [0.0492, 0.0515], [0.0352, 0.0376]).

The subjective preference values of decision-making group to plan $A_i, i = \overline{1,5}$ are
$q_1 = (0.2, 0.3, 0.3, 0.4)$, $q_2 = (0.2, 0.4, 0.4, 0.5)$, $q_3 = (0.1, 0.2, 0.3, 0.4)$,
$q_4 = (0.1, 0.2, 0.3, 0.4)$, $q_5 = (0.2, 0.3, 0.4, 0.5)$.

The relative approach degree of grey TOPSIS method is
$C = (C_1, C_2, C_3, C_4, C_5) = (0.5718, 0.8550, 0.4805, 0.2247, 0.5841)$.

**Table 1.** Decision matrix

| Index Plan | $G_1$ | $G_2$ | $G_3$ | $G_4$ | $G_5$ |
|---|---|---|---|---|---|
| $A_1$ | (3610, [0.2, 0.4]) | (490, [0.3, 0.5]) | ([465,485], [0.2,0.25]) | (4890, [0.4, 0.6]) | ([850,950], [0.2, 0.4]) |
| $A_2$ | ([3540, 3640], [0.3, 0.5]) | (520, [0.2, 0.4]) | ([480,490], [0.5,0.6]) | ([4680,4790], [0.2, 0.4]) | ([800,900], [0.4, 0.6]) |
| $A_3$ | (3700, [0.1,0.3]) | ([460,500], [0.3, 0.5]) | (470, [0.2,0.3]) | ([4600,4720], [0.3, 0.5]) | ([700,800], [0.5, 0.7]) |
| $A_4$ | ([3730,3830], [0.2, 0.4]) | (470, [0.1, 0.3]) | ([460,475], [0.3,0.5]) | (4715, [0.2, 0.3]) | ([700,750], [0.4, 0.6]) |
| $A_5$ | (3690, [0.3, 0.4]) | ([490,530], [0.2,0.4]) | ([470,485], [0.4,0.6]) | ([4790,4850], [0.2, 0.3]) | ([750,850], [0.4, 0.5]) |

**Table 1**. Decision matrix (continued)

| Index Plan | $G_6$ | $G_7$ | $G_8$ | $G_9$ |
|---|---|---|---|---|
| $A_1$ | (very high, [0.3, 0.5]) | (rather high, [0.4,0.6]) | ([a little high, rather high], [0.5,0.7]) | (rather low, [0.3,0.5]) |
| $A_2$ | ([rather high, very high], [0.3, 0.5]) | (high, [0.6,0.7]) | ([high, very high], [0.2,0.4]) | ([low, rather low], [0.5,0.6]) |
| $A_3$ | ([rather high, high], [0.3, 0.6]) | ([a little high, high], [0.4,0.7]) | (high, [0.3,0.5]) | ([very low, rather low], [0.2,0.4]) |
| $A_4$ | (general, [0.2,0.5]) | (a little high, [0.3,0.6]) | ([general, rather high], [0.4,0.6]) | ([rather low, a little low], [0.3,0.5]) |
| $A_5$ | (rather high, [0.3,0.6]) | ([rather high, rery high ], [0.2,0.5]) | ([rather high, high], [0.3,0.5]) | ([very low, low], [0.2,0.4]) |

Therefore, we obtain the rank such as

   Plan 2 ≻ plans 5 ≻ plan 1 ≻ plan 3 ≻ plan 4

Then, we calculate the degree of grey relation relative approach with the preference coefficients. For the preference coefficients $\theta_+ = \theta_- = 0.5$, we obtained

  $C' = (C'_1, C'_2, C'_3, C'_4, C'_5) = (0.4869, 0.5534, 0.4760, 0.4501, 0.5231)$

and we obtain the plan rank such as

   Plan 2 ≻ plans 5 ≻ plan 1 ≻ plan 3 ≻ plan 4

The relative membership degree between the given plans and the positive ideal plan is

$u = (u_1, u_2, u_3, u_4, u_5) = (0.4737, 0.6055, 0.4521, 0.4012, 0.5461)$

and we obtain the rank such as

   Plan 2 ≻ plans 5 ≻ plan 1 ≻ plan 3 ≻ plan 4.

The grey incidence relative approach degree using the maximum entropy estimation is

  $C''_i = (C''_1, C''_2, C''_3, C''_4, C''_5) = (0.8083, 0.9327, 0.7969, 0.7821, 0.8729)$

and the corresponding rank is

   Plan 2 ≻ plans 5 ≻ plan 1 ≻ plan 3 ≻ plan 4.

The final rank determined by the weighted Borda method is

   Plan 2 ≻ plans 5 ≻ plan 1 ≻ plan 3 ≻ plan 4.

# 6. Conclusion

 We proposed a generalized fuzzy grey decision making method, which takes into consideration of the grey degree of the weight and the attribute value at the same time, for the MADM where attributes have the generalized super mixed-type values given by real number, interval value, linguistic value and uncertain linguistic value. First, based on the attribute values given by decision maker, we obtained the grey degree corresponding to fuzzy part of the grey fuzzy comprehensive decision matrix, and then obtained a generalized decision matrix composed of four-dimensional vector dealing with the grey part of decision matrix by using the information sufficiency degree. Second, we obtained four ranks by four methods of plan evaluation such as the evaluation by the relative approach degree of grey TOPSIS, the evaluation by the relative approach degree of grey incidence, the evaluation by the relative membership degree of grey incidence and the evaluation by the grey relation relative approach degree using the maximum entropy estimation. Finally, using the ranks obtained from the above four methods, the final rank are determined by the weighted Borda method.